\documentclass[runningheads]{llncs}
\usepackage[T1]{fontenc}
\usepackage{amsmath}
\usepackage{amssymb}
\usepackage{amsfonts}
\usepackage{pifont}
\usepackage{algorithmic}
\usepackage[ruled, vlined, linesnumbered]{algorithm2e}
\usepackage{todonotes}
\usepackage{multirow}
\usepackage{soul}
\usepackage{subcaption}
\usepackage{hyperref}
\usepackage{graphicx}
\usepackage{appendix}
\newcommand{\nop}[1]{}

\newcommand{\eg}{\emph{e.g., }}
\newcommand{\ie}{\emph{i.e., }}
 
\DeclareMathOperator*{\argmin}{arg\,min} 

\usepackage[misc]{ifsym}
\usepackage{color}

\begin{document}
\tocauthor
\toctitle
\title{SECLEDS: Sequence Clustering in Evolving Data Streams via Multiple Medoids and Medoid Voting}
\titlerunning{SECLEDS: Sequence Clustering in Evolving Streams}
%\author{Anonymous author(s)}
\author{Azqa Nadeem(\Letter) \and
Sicco Verwer}%\orcidID{0000-0002-3950-0542}
\authorrunning{A. Nadeem, S. Verwer}
\institute{Delft University of Technology, The Netherlands\\
\email{\{azqa.nadeem,s.e.verwer\}@tudelft.nl}}
\maketitle              % typeset the header of the contribution
\begin{abstract}
Sequence clustering in a streaming environment is challenging because it is computationally expensive, and the sequences may evolve over time. 
K-medoids or Partitioning Around Medoids (PAM) is commonly used to cluster sequences since it supports alignment-based distances, and the \texttt{k}-centers being actual data items helps with cluster interpretability.  
However, offline k-medoids has no support for concept drift, while also being prohibitively expensive for clustering data streams. 
We therefore propose SECLEDS, a streaming variant of the k-medoids algorithm \textit{with constant memory footprint}. SECLEDS has two unique properties: i) it uses multiple medoids per cluster, producing stable high-quality clusters, and ii) it handles concept drift using an intuitive \textit{Medoid Voting scheme} for approximating cluster distances. 
Unlike existing adaptive algorithms that create new clusters for new concepts, SECLEDS follows a fundamentally different approach, where the clusters themselves evolve with an evolving stream.
Using real and synthetic datasets, we empirically demonstrate that SECLEDS produces high-quality clusters regardless of drift, stream size, data dimensionality, and number of clusters.
We compare against three popular stream and batch clustering algorithms. The state-of-the-art BanditPAM is used as an offline benchmark. 
SECLEDS achieves comparable F1 score to BanditPAM while reducing the number of required distance computations by 83.7\%.
Importantly, SECLEDS outperforms all baselines by 138.7\% when the stream contains drift. 
We also cluster real network traffic, and provide evidence that SECLEDS can support network bandwidths of up to 1.08 Gbps while using the (expensive) dynamic time warping distance.

\end{abstract}

\section{Introduction}

Stream clustering is the problem of clustering a potentially unbounded stream of items in a single pass, where the items arrive sequentially without any particular order, \eg network traffic, financial transactions, and sensor data. Stream clustering algorithms must have low memory overhead, be computationally efficient, and robust to concept drift, \ie evolving data distributions \cite{silva2013data}.
Maintaining high cluster quality in a fully online setting is extremely difficult. Therefore, hybrid online-offline algorithms are popular among existing approaches, \eg CluStream \cite{aggarwal2003framework}, StreamKM++ \cite{ackermann2012streamkm}, DenStream \cite{cao2006density}, BIRCH \cite{zhang1997birch}. 
These algorithms have an online component that summarizes the data stream, and an offline component that periodically uses that information to create the final clusters. There also exist algorithms that store part of the stream for handling outliers, \eg BOCEDS \cite{islam2019buffer}, MDSC \cite{fahy2019finding}. 
Existing stream clustering algorithms handle concept drift by having variable number of clusters: they add new clusters for newly observed behavior and discard clusters that contain too many old data items. 
This leads to higher memory requirements for managing buffers and intermediate solutions.
Batch clustering algorithms can also be used in a streaming setting by considering a batch size of one, \eg Minibatch k-means \cite{sculley2010web}. However, they start to under-perform when the stream contains drift. 

In recent years, sequential data has increasingly become popular because of the powerful insights that it provides regarding behavior analytics \cite{boeva2019modeling}, \eg for attacker strategy profiling \cite{nadeem2021alert}, fraud detection \cite{guo2018learning}, human activity recognition \cite{cook2013activity}. 
Clustering sequences in an offline setting is challenging in itself because sequences are often out-of-sync, requiring expensive alignment-based distance measures, which are often not supported by many clustering algorithms. 
K-medoids or Partitioning Around Medoids (PAM) has often been used to cluster sequences because the \texttt{k}-centers are represented by actual data items, called medoids or prototypes \cite{wang2019k,ushakov2021near}. This has multiple benefits: i) it makes the cluster interpretation simpler; ii) it enables the use of non-metric distances such as dynamic time warping (DTW); and iii) it allows to estimate exact storage requirements based on the \texttt{k}-fixed clusters. 
Although the state-of-the-art offline k-medoids algorithms, \ie FastPAM1 \cite{schubert2019faster} and BanditPAM \cite{tiwari2020banditpam} have reduced the runtime complexity to $\mathcal{O}(nlogn)$, they are still not efficient enough to be used in streaming settings, and the cluster quality will degrade over time as the stream evolves. 
To the best of our knowledge, there exists no streaming version of the k-medoids algorithm that can efficiently cluster sequential data.

\begin{figure}[t]
    \centering
    \includegraphics[width=\linewidth]{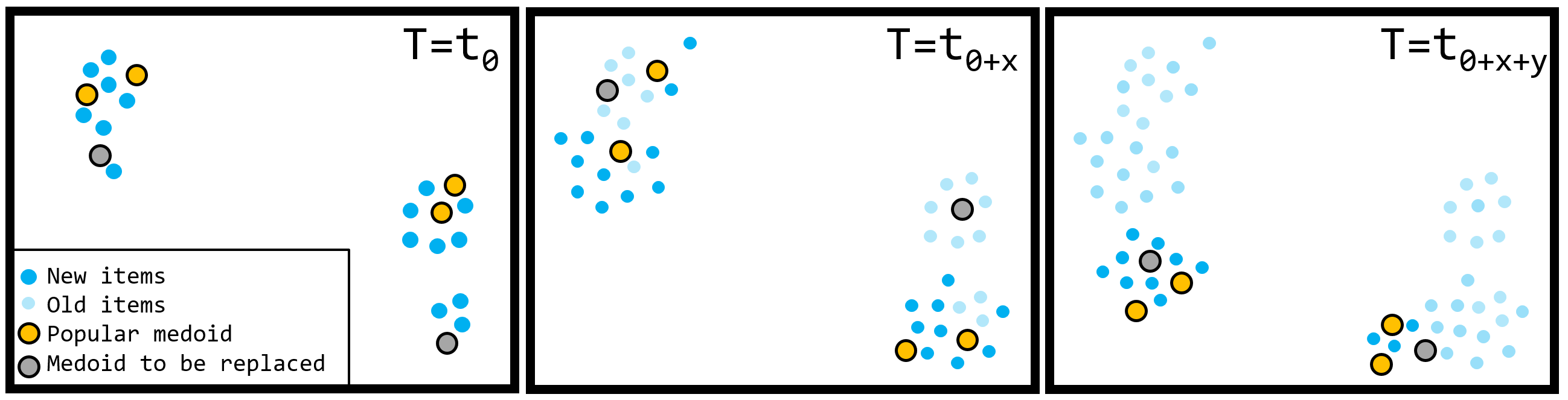}
    \caption{An illustration of SECLEDS' clusters following an evolving data stream. The medoids close to recent data gain more votes, while the medoids with the least votes are replaced with new data items from the stream. In effect, the \texttt{k}-clusters handle concept drift by capturing different concepts in the stream at different time steps.}
    \label{fig:decay-over-time}
\end{figure}

% answer
\textbf{Contributions.} In this paper, we propose \textit{SECLEDS}, a lightweight streaming version of the k-medoids algorithm with \textit{constant memory footprint}. 
SECLEDS has two unique properties: 
Firstly, it uses \texttt{p}-medoids per cluster to maintain stable high-quality clusters. Note the difference from IMMFC \cite{wang2014incremental}, which uses the information of multiple medoids in independent sub-solutions to select the final medoids. We initialize the \texttt{p}-medoids using a non-uniform sampling strategy similar to k-means++.
Secondly, a Medoid Voting scheme is used to estimate a cluster's center of mass. The offline k-medoids has a \textsc{Swap} step that tests each point in a cluster to determine the next medoid. SECLEDS cannot do this because it does not store any part of the stream. Instead, it maintains votes for each medoid that estimate how representative (valuable) it is given the data seen so far. A user-supplied decay factor enables SECLEDS to slowly forget the votes regarding past data. The least representative medoids are then replaced with new data items. This way, rather than creating new clusters for new concepts, the \texttt{k}-clusters themselves evolve with the data stream. Figure \ref{fig:decay-over-time} shows how the clusters follow a data stream as it evolves. Thus, the \texttt{k}-clusters represent different concepts in the stream at different time steps. 
SECLEDS addresses the following real-world constraints:
\begin{enumerate}
    \item[I.] A runtime efficient medoid-based clustering algorithm with a fixed memory footprint that can handle high-bandwidth data streams,
    \item[II.] An algorithm that produces high-quality clusters in a streaming environment while being able to deal with concept drift,
    \item[III.] Accurate sequence clusters using alignment-based distances, and
    \item[IV.] Minimal parameter settings to support ease-of-use.
\end{enumerate}

\textbf{Empirical results. } We experiment on several real and synthetic data streams that contain 2D points and univariate sequences. We empirically demonstrate that SECLEDS produces high-quality clusters regardless of drift, stream size, and number of clusters. We use the following state-of-the-art and popular clustering algorithms as baselines: a) Streaming: CluStream, StreamKM++; b) Batch: Minibatch k-means; c) Offline: BanditPAM. Particularly, BanditPAM is used as a benchmark for the best achievable clustering on a static dataset. 
The results show that i) SECLEDS achieves comparable F1 score to BanditPAM, while reducing the required number of distance computations by 83.7\%; ii) SECLEDS outperforms all baselines by 138.7\% when the stream contains drift; iii) SECLEDS is faster than BanditPAM and CluStream on most clustering tasks.

We also discuss a use-case from the network security domain where network traffic is often randomly sampled to keep the storage requirements within a predefined budget. Consequently, temporal patterns in the network traffic are lost that could have been useful for downstream tasks, \eg behavior analytics. We propose a smarter sampling technique that uses medoid-based stream clustering (SECLEDS) to summarize the network traffic: SECLEDS clusters sequences of network traffic and periodically stores the medoids of each cluster, thus reducing the storage needs while preserving temporal patterns in the data. By clustering real-world network traffic, we provide evidence that SECLEDS (and SECLEDS-dtw) can support network bandwidths of up to 2.79 Gbps (and 1.08 Gbps), respectively. 
We release SECLEDS as open-source\footnote{SECLEDS: \url{https://github.com/tudelft-cda-lab/SECLEDS}}.

\section{Preliminaries} 
\textbf{Stream.} Given a sensor that receives an unbounded stream of multi-dimensional data points $X  = \{x_1, x_2, \dots\}$ with dimensionality $d$, arriving at time steps $T = \{t_1, t_2, \dots\}$, 
a sequential data stream is defined as $S = \{s_1,\dots s_n,\dots \}$, where $s_i$ is a time window $w$ over $X$ such that $s_i = \{x_i, x_{i+1}, \dots x_{i+w}\}$, and $y_i$ is its associated class label. Traditional point clustering considers $w = 1$, while for sequence clustering, we consider $w > 1$. 
%The data point $x$ is assumed to have $d$ dimensions.
We use two configurations, \ie $d=2, w=1$ (2D point clustering) and $d=1, w=100$ (univariate sequence clustering). A case of bivariate variable length sequences is given in appendix.

\textbf{Concept drift.} Real-world data streams often change unexpectedly over time. This shift alters the statistical properties of their underlying distribution. In machine learning, this is called concept drift \cite{lu2014concept,vzliobaite2016overview}. Concept drift is typically categorized into four types \cite{lu2018learning}: (i) \textit{Sudden drift} where a new concept arises abruptly; (ii) \textit{Gradual drift} where an old concept is slowly replaced by a new one; (iii) \textit{Incremental drift} where a concept incrementally turns into another one; and (iv) \textit{Recurring concepts} are old concepts that reappear from time to time. 
Years of research has gone into developing \textit{concept drift detectors} that either monitor the underlying data distribution, error rate or perform hypothesis testing to trigger model retraining \cite{lu2018learning,barros2018large}. Typical stream clustering algorithms handle concept drift by introducing new clusters for new concepts, and discarding old irrelevant clusters \cite{hyde2017fully}. Although intuitively appealing, this requires user-supplied parameters that define what `new' means. 

\section{SECLEDS: Sequence Clustering in Evolving Streams}

SECLEDS is a lightweight streaming variant of the classical k-medoids (PAM) algorithm. 
To support high bandwidth data streams, SECLEDS does not store any part of the stream in memory --- it receives an item, assigns it to one of the \texttt{k}-clusters, and then discards it. This way, SECLEDS has a guaranteed constant memory footprint [\textbf{I}]. 
However, this requirement cannot be achieved using the offline \textsc{Build} and \textsc{Swap} steps of PAM. Instead, SECLEDS performs a non-uniform sampling (similar to k-means++) on an initial batch of the stream to initialize the medoids. It also makes use of \textit{multiple medoids per cluster} to provide a stable cluster definition in a streaming setting where noise and concept drift are common properties [\textbf{II}]. 

The efficiency of a chosen distance measure is usually the primary performance bottleneck in sequence clustering. Thus, minimizing the number of distance computations is key to scaling SECLEDS to large data streams. This is achieved by introducing a \textit{Medoid Voting scheme} whose purpose is twofold: (i) it determines the center of mass of a cluster, thus is able to estimate how representative a medoid is given the data seen so far; and (ii) by using this information, old irrelevant medoids are replaced by new ones that are located near recent data. Hence, better medoids can be found without having to perform additional distance computations [\textbf{I}]. This also allows SECLEDS to support robust but computationally expensive distance measures specifically meant for sequential data, \eg dynamic time warping [\textbf{III}]. Finally, SECLEDS handles concept drift by regularly \textit{forgetting} past data and occupying newer regions/concepts in the data stream. This is achieved by applying exponential decay $\lambda$ to the medoid votes at each time step [\textbf{II}]. 

SECLEDS has a modular implementation in Python. The \texttt{k}-clusters, \texttt{p}-medoids per cluster, and decay rate $\lambda$ are the only three user-supplied parameters needed for the algorithm, making it useful for exploratory data analysis [\textbf{IV}]. We believe these parameters are easier to tune compared to many radius- or density-based hyperparameters in existing clustering algorithms, which require a deeper understanding of the data distribution in advance.

\begin{figure}[t]
    \centering
    \includegraphics[width=0.9\linewidth]{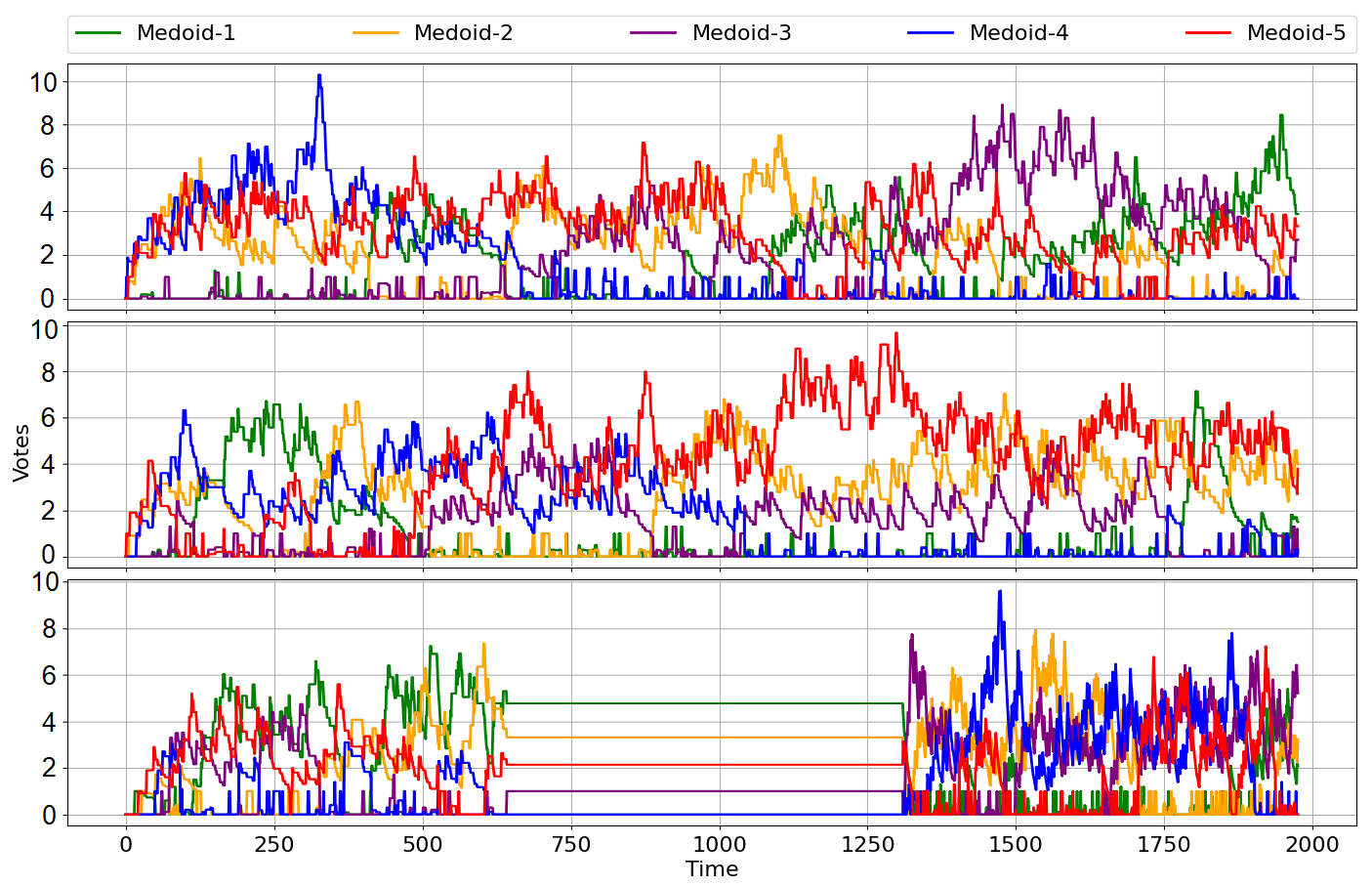}
    \caption{The effect of cluster initialization and concept drift on medoid selection of a cluster, where $p=5$. (Top): Given a uniformly distributed stream, every medoid becomes popular at some point. (Middle): For an incrementally drifted stream, the medoid close to drifted data becomes popular. (Bottom): For a class-ordered stream, all clusters start from one class, until one medoid migrates to the correct class. In this case, the correct class is observed from $t_{1300}$. Medoid-3 migrates first and becomes popular, while medoid-1 migrates last.}
    \label{fig:votesOT}
\end{figure}

\subsection{Stable cluster definition via multiple medoids}
A new data item $s$ is assigned to a cluster $cid$ with the least average distance to its medoids. With multiple medoids per cluster, this provides a robust cluster assignment.
Additionally, the medoid voting scheme encourages the medoids to represent different sections of a class so they can gain votes: if they are too close together, some of them do not receive votes and get replaced eventually. It also ensures that outliers are quickly replaced because of fewer votes.

Concept drift and cluster initialization determine how the medoids behave. Figure \ref{fig:votesOT} illustrates three scenarios with varying medoid behavior for a single cluster as a function of votes gained over time:
(a) Assuming no concept drift, when the stream is roughly evenly shuffled, all the medoids receive uniform votes on average. This is because all medoids are close to parts of the stream at different time steps. At a specific time step, the medoid close to the most amount of recent data becomes popular. The top figure shows that each medoid becomes popular at some point in the stream, indicating that the medoids represent different sections of the underlying class. 
(b) When the stream incrementally drifts, the cluster follows the evolving stream by replacing the least popular medoids with recent data items from the stream. Since the new medoids are now closer to new data, they gain more votes and become popular. This has roughly the same effect as the first case. 
(c) When the data arrives one class at a time, all clusters are initialized in a single class. 
As data from a new class appears, one medoid from the closest cluster migrates to it and starts gaining votes. Over time, the older popular medoids lose their votes because of exponential decay, and eventually migrate to the new class. This is shown from $t_{1300}$ onward in the bottom figure, highlighting the importance of multiple medoids in noisy streams.

\subsection{Center of mass estimation} 
The voting scheme provides an estimate of a cluster's center of mass by assigning more votes to recently observed data in the stream $S$, while exponential decay helps to forget votes regarding older data. Without decay, older clusters with popular medoids never evolve. Thus, these properties help to replace irrelevant medoids, \eg those that are located i) close to the least amount of recent data, or ii) in a region where new data no longer arrives.
Note that we only apply exponential decay to the \textit{most recently updated cluster}, so that we do not forget valuable information about other clusters while the data from this class arrives.

\section{The SECLEDS Algorithm}

SECLEDS has three modules: an initialization module (\textsc{Init}), an assignment module (\textsc{Assign}), and an update module (\textsc{Update}). The task is to assign each item in $S$ to one of the \texttt{k}-clusters.  
SECLEDS maintains and updates a model of the stream seen so far in the form of \texttt{k}-clusters, $\mathcal{C} = \{ C_1, \dots C_k \}$. For clarity, we use $t$ to denote the clusters at time $t$. These superscripts are removed from Algorithm \ref{alg:SECLEDS}. Each cluster is represented by a set of \texttt{p}-medoids and their votes, \ie for $1 \leq i \leq k$, $C_i^t = \{(m^t_{i,1}, v^t_{i,1}) \ldots (m^t_{i,p}, v_{i,p}^t) \}$, where for $1 \leq j \leq p$: $m^t_{i,j} \in S$ is the $j^{th}$ medoid of the $i^{th}$ cluster at time $t$ having $v_{i,j}^t \in \mathbb{R}$ votes.

\textbf{\textsc{Init}.} A batch $\mathbb{B}$ from the start of $S$ is used to initialize the clusters. The batch can be small but enough to select $k\cdot p$ medoids. In the experiments, we used a batch size of ($1.5\cdot k\cdot p$). We use a non-uniform sampling strategy, similar to the Lloyd's algorithm \cite{lloyd1982least}, to select the primary medoid of each cluster: SECLEDS selects the first medoid of the first cluster ($m_{1,1}^1$) arbitrarily from the batch. Another $k-1$ medoids are sampled with a probability proportional to the squared distance between $m_{1,1}^1$ and other items in $\mathbb{B}$. This initializes the primary medoid of each cluster. The other \texttt{p-1} medoids for each cluster $C_i^1$ are the items in $\mathbb{B}$ that are closest to its primary medoid $m_{i,1}^1$. This way, the medoids maintain cluster separation by reducing the risk of medoids from multiple clusters overlapping each other. All medoids start with 0 votes.

\textbf{\textsc{Assign} and \textsc{Update}.} With the clusters initialized, the stream processing begins. \textsc{Assign} and \textsc{Update} are called for each item in $S$. 
\textsc{Assign} has 3 steps: (i) An incoming item $s$ at time $t$ is assigned to the cluster $C_{cid}^t$ for which its previous medoids $C_{cid}^{t-1}$ have the least average distance to $s$, formally defined in Eq. (\ref{frac:assignment}) for any given distance function $d(.,.)$. (ii) The closest medoid to $s$ receives a vote, while exponential decay $\lambda$ is applied to the other medoids \ie for all $1 \leq j \leq p$ and $j' \neq j$: $v_{cid,j}^t = (v_{cid,j}^{t-1}+1)$ if $j = \argmin_j d(s, m_{cid,j}^{t-1})$, otherwise $v_{cid,j'}^t = v_{cid,j'}^{t-1}\cdot(1-\lambda)$. This way, the medoids maintain an estimate of their centers of mass without storing any part of the stream. (iii) The votes of all other clusters remain the same, \ie $v_{i,j}^t = v_{i,j}^{t-1}$ for all $i \not= cid$ and $1 \leq j \leq p$.

\begin{equation}
    C_{cid}^{t} = \argmin_{1 \leq cid \leq k} \frac{\sum_{j=1}^{p} d(s, m_{cid,j}^{t-1})}{p}
\label{frac:assignment}
\end{equation}

At every time step $t$, the new data item $s$ is promoted to be a medoid of $C_{cid}^t$: the medoid having the least votes which is not the newest medoid is replaced by $s$, \ie $\{m_{cid,j}^t = s,\ v_{cid,j}^t = 0\}$  where $j = \argmin_{j} v_{cid,j}^{t-1}$ and $m_{cid,j}^{t-1} \neq \eta_{cid}^t$, where $\eta_{cid}^t$ keeps track of the newest medoid of cluster $cid$ at time $t$. Inspired by Tabu search \cite{glover1986future}, including $\eta_{cid}^t$ ensures that the most-recently updated medoid is not selected to be replaced each time. Tabu search is a local search meta-heuristic that selects which values to change except for the last $\delta$ ones ($\delta=1$ in this case).

\begin{algorithm}[t]
\DontPrintSemicolon
\KwIn{Data stream, nclusters, nprototypes: $S,\ k,\ p$}
\SetKwBlock{Begin}{function}{end function}
\Begin($\textsc{SECLEDS} {(} S,k,p {)}$)
{   
     $b \gets 1.5 \cdot k \cdot p$\;
     $\mathbb{B}$ $\gets$ Collect b items from $S$\;
     $\mathcal{C}$ $\gets$ \textsc{Init}($\mathbb{B}$, $k$, $p$) \ \ \ \ \ \ \ \ \ \ \ \ \ \ \ \ \ \ \ \ \ \ \ \ \ \ \ \ \ \ \ \ \ \ \ \ \ \ \ \ \ \ \ \ \ \ \ \ \ \ \ \ \ \ \ \ \  \tcp*[f]{\textsc{Init}}\;
    \ForAll{$s$ in $S[b:]$} 
       {  
        $cid\ \gets\ \argmin_{1 \leq cid \leq k} \frac{1}{p}\cdot \sum_{j=1}^{p} d(s, m_{cid,j})$ \ \ \ \ \ \ \ \ \ \ \ \ \ \ \ \ \ \ \ \ \ \ \ \  \tcp*[f]{\textsc{Assign}}\; 
        $j\ \gets\ \argmin_j d(s, m_{cid,j})$ for all $ 1 \leq j \leq p$\;
        $v_{cid,j} \gets (v_{cid,j}+1)$, $v_{cid,j'} \gets v_{cid,j'}\cdot (1-\lambda)$ for $j' \neq j$\;
        $j\ \gets\ \argmin_j v_{cid,j}$ where $m_{cid,j} \neq \eta_{cid}$ for all $1 \leq j \leq p$ \ \ \  \tcp*[f]{\textsc{Update}}\;
        $m_{cid,j} \gets s,\ v_{cid,j} \gets 0$\; 
        \textbf{yield} $cid$\;
        }
}   
\Begin($\textsc{Init} {(} \mathbb{B},k,p {)}$)
{
    Choose $m_{1,1} \in \mathbb{B}$ arbitrarily. Let $C_1 \gets \{(m_{1,1},0)\}$\;
    \For{$i\gets 2 \dots k$}
    {
        Choose $m_{i,1} \in \mathbb{B}$ with probability $d(m_{i,1}, m_{1,1})^2$, $m_{i,1}\neq m_{1,1}$\;
        Let $C_i \gets \{(m_{i,1},0)\}$\;
    }
    \For{$i\gets1 \dots k$}
    {
        $dist$ $\gets$ $d(b, m_{i,1})$ for all $b \in \mathbb{B}$ and $b\neq m_{i,1}$\;
       Choose $\{m_{i,2} \dots m_{i,p}\}$ having smallest values in $dist$\;
       Update $C_i \gets \{(m_{i,1},0) \dots (m_{i,p},0)\}$\;
       }
    \Return{$\{C_1,\dots,C_k\}$}
}
\caption{SECLEDS for clustering sequences in evolving streams}
\label{alg:SECLEDS}
\end{algorithm}

\paragraph{\textbf{Time complexity.}} Given \texttt{k} clusters, \texttt{p} medoids, \texttt{b} batch size, and \texttt{n} items in the stream, SECLEDS has a time complexity of $\mathcal{O}(\texttt{n})$: SECLEDS selects the first medoid at random from the initial batch, and then performs \texttt{b} distance computations to find the other \texttt{k-1} primary medoids. The rest of the \texttt{k(p-1)} medoids are also selected using the same distance information. In total, this requires $\mathcal{O}(\texttt{kb})$ distance computations. For every $s \in S$, SECLEDS computes the average distance to each cluster, which requires \texttt{kp} distance computations. Over an entire run, this gives \texttt{nkp} distance computations. In the \textsc{Update} module, SECLEDS reallocates medoid votes without any distance computations, making the runtime negligible. Since \texttt{k} and \texttt{p} are small user-supplied parameters, the overall runtime complexity is $\mathcal{O}(\texttt{n})$. 

\paragraph{\textbf{Space complexity.}} After initialization, SECLEDS only stores the \texttt{p} medoids and their votes for the \texttt{k} clusters. Since these are (small) user-defined parameters, the space complexity of SECLEDS is $\mathcal{O}(\texttt{1})$.

\section{Experimental Setup}

\paragraph{\textbf{Datasets.}}
We use three synthetic and a real dataset containing 2D points and univariate sequences, see Table \ref{tab:dataset-properties}. The data generation process is given in appendix. The synthetic datasets are released in the SECLEDS code repository. 

\textbf{Blobs:} The blob dataset was created using \texttt{scikit-learn} \cite{pedregosa2011scikit}. The dataset contains $n=100,000$ two-dimensional points ($d=2$), equally distributed in $k=10$ classes, with varying standard deviations.

\textbf{Sine-curve:} A sine-curve generator was used to create $k=4$ synthetic univariate sine curves of length $1,250,000$ each, using varying \textit{frequency}, \textit{phase} and \textit{error} (see appendix). Each curve is partitioned using a non-overlapping window of length $w=100$ to obtain the experimental dataset. In total, $n=50,000$ curves are obtained, equally divided across $k=4$ classes.

\textbf{Sine-curve-drifted:} Incremental concept drift is added to the Sine-curve dataset by shifting the phase of each curve by a factor of $(drift \cdot c\_id)$, where $c\_id$ is the curve index in the stream, and $drift=0.05$. Note that adding drift to the frequency of the sine curves produces similar results.

\textbf{CTU13-9:} CTU13 \cite{garcia2014empirical} is an open source dataset composed of network traffic (netflows) coming from real botnet-infected hosts and normal hosts. We use scenario-9, containing 10 bot-infected hosts and 6 benign hosts. A total of $2,087,509$ (normal and botnet) netflows were captured over 5 hours and 37 minutes. We obtain $n=213,386$ univariate sequences of length $w=100$ using a sliding window model \cite{zubarouglu2021data} with $step\_size=1$ (see appendix).  

We use Euclidean distance for all datasets. For the Sine-curve and network traffic datasets, we additionally use Dynamic time warping (DTW). Note that Euclidean distance can only be used with fixed-length sequences and often produces less accurate results compared to DTW \cite{wang2013experimental,guijo2020time}.

\begin{table}[t]
\centering
\caption{Summary of experimental datasets.}
\label{tab:dataset-properties}
\resizebox{\columnwidth}{!}{%
\begin{tabular}{lccccc}
\hline
\textbf{Dataset} & \textbf{ Type } & \textbf{Drift} &\textbf{Stream size (n)} & \textbf{Clusters (k)} & \textbf{Dimensions (d,w)}\\ \hline
\textbf{Blobs} & Synthetic& No &100,000 & 10 & (2,1) \\ \hline
\textbf{Sine-curve} & Synthetic & No &50,000 & 4 & (1,100) \\ \hline
\textbf{Sine-curve-drifted} & Synthetic & Yes &50,000 & 4 & (1,100) \\ \hline
\textbf{CTU13-9} & Real-world & Yes &213,386 & 2 & (1,100) \\ \hline
\end{tabular}
}
\end{table}

\paragraph{\textbf{Stream configuration.}} A data stream $S$ of size $n$ is constructed from a chosen experimental dataset. For each experiment, the clustering task is executed $trials$-times, randomly shuffling the stream each time, to make the results data-order-invariant. A clustering task invokes SECLEDS and the baselines such that each algorithm observes the exact same order of data arrival. In this paper, we set $trials=10$, unless otherwise reported. All experiments are run on Intel Xeon E5620 quad-core processor with 74 GB RAM.

\paragraph{\textbf{Evaluation. }}
We use two metrics for performance evaluation: i) \textit{runtime} to cluster a stream size of $n$; ii) \textit{F1 score} computed from the pairwise co-occurrences of items in the stream using Eq. (\ref{eq:f1}), as originally defined in \cite{manning2010introduction}.

\begin{align}
    & eval(a,b) = \begin{cases}
y_a = y_b \land C_x = C_y, &\text{true positive}\\
y_a = y_b \land C_x \neq C_y, &\text{false negative}\\
y_a \neq y_b \land C_x = C_y, &\text{false positive}\\
y_a \neq y_b \land C_x \neq C_y, &\text{true negative}\\
\end{cases}
\label{eq:f1}
\end{align}

 where $y_a$ and $y_b$ are labels of items $a$ and $b$ that are placed in clusters $C_x$ and $C_y$. Since clusters do not have pre-defined labels, data from one class may be assigned to arbitrary clusters in different runs. Thus, instead of looking at the predicted label, we measure F1 using the pairwise co-occurrences of true labels.

\paragraph{\textbf{Baselines.}} We compare SECLEDS with state-of-the-art open-source partition-based clustering algorithms with k-fixed clusters: 
a) Streaming: CluStream, StreamKM++; b) Batching: MiniBatch k-means; c) Offline: BanditPAM. 
MiniBatch k-means and StreamKM++ are online versions of k-means, while CluStream is an adaptive, online-offline algorithm. 
BanditPAM (v1.0.5) is used as a benchmark for the best achievable clustering on a static dataset. 
We set time\_window=1, max\_micro\_clusters=$k\cdot p$, halflife=0.5 for CluStream; chunk\_size=1, halflife=0.5 for StreamKM++; batch\_size=1, max\_iter=1 for Minibatch k-means.

\begin{figure}[t]
    \centering
    \includegraphics[width=\linewidth]{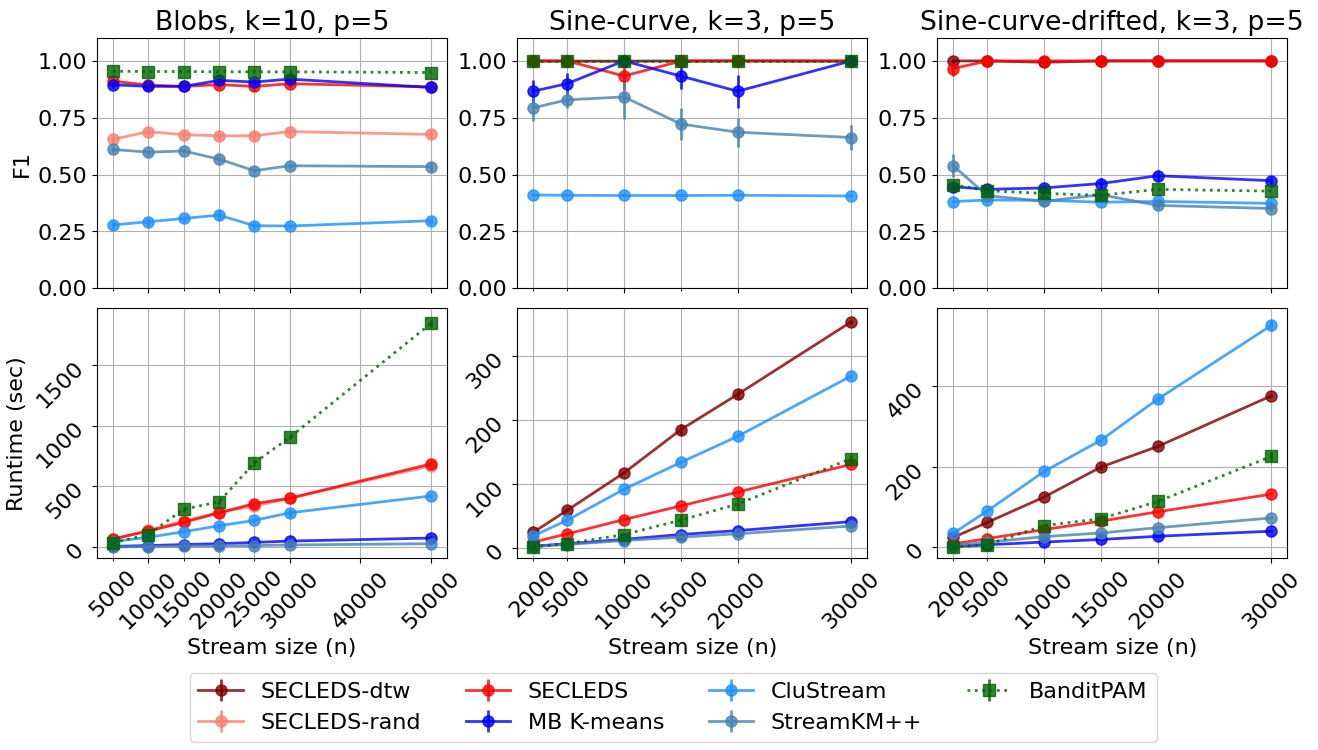}
    \caption{Clustering Blobs and Sine-curves: SECLEDS's runtime grows approximately linearly with stream size, while maintaining competitive F1 score with the best-performing baselines, \ie BanditPAM and Minibatch k-means. SECLEDS consistently performs better than all baselines in the presence of concept drift.}
    \label{fig:results-overall}
\end{figure}

\section{Empirical Results}

\textit{Key findings. } In this section, we empirically demonstrate the following results:
\begin{enumerate}
    \item SECLEDS produces high-quality clusters, regardless of concept drift, stream size $n$, data dimensionality $(d,w)$, and number of clusters $k$. SECLEDS shows competitive F1 compared to the best performing baseline (BanditPAM), while reducing the number of required distance computations by 83.7\%.
    \item SECLEDS outperforms \textit{all baselines} by 138.7\% when the stream contains concept drift. SECLEDS outperforms the \textit{best-performing streaming baseline} by 58.2\% on Blobs, 33.3\% on Sine-curve, and 143.7\% on Sine-curve-drifted.
    \item SECLEDS-dtw clusters $\sim$5.5h of network traffic in just 8\% of the time. Thus, it can handle networks with bandwidths of up to 1.08 Gbps, which is significantly higher than the requirements of a typical enterprise network.  
\end{enumerate}

\paragraph{\textbf{Point vs. sequence clustering. }}\label{sec:point-vs-seq-clustering}

We use Blobs with $k=10$ on stream sizes $n=(5000,\dots 50,000)$, and Sine-curve with $k=3$ on stream sizes $n=(2000,\dots 30,000)$. For both, we set $p=5$, $\lambda=0.1$, $trials=10$. The mean and standard deviation of the F1 scores and runtimes are given in Figure \ref{fig:results-overall}. The benchmark (BanditPAM) achieves a mean F1 of 0.95 and 1.0 for Blobs and Sine-curve, respectively. 

SECLEDS outperforms both CluStream and StreamKM++ on the Blobs dataset, and additionally outperforms Minibatch k-means on the Sine-curve dataset. 
Minibatch k-means performs exceptionally well on point clustering, but loses its edge on sequence clustering. This is because the centroids are computed by collapsing temporally-linked dimensions into single values that do not adequately represent the sequences.
An improvement in F1 score is observed for CluStream and StreamKM++ on the higher dimensional Sine-curve dataset because of fewer clusters ($k=10$ vs. $k=3$). 

We also compare the effect of euclidean and dynamic time warping distance on the Sine-curve dataset. Although, they both produce equivalent results, it must be noted that euclidean distance only works with fixed-length sequences. An example of SECLEDS-dtw on clustering bivariate sequences $d=2,w=$(min:15, max:121) from \textit{UJI Pen Characters} \cite{Dua2019} is given in the appendix. 

\paragraph{\textbf{Initialization quality. }} Stream clustering algorithms are greatly impacted by the quality of cluster initialization. To test this, we compare SECLEDS against SECLEDS-rand (initialized with randomly selected medoids from the initial batch $\mathbb{B}$). 
Evidently, the clusters take a long time to converge, regardless of the stream size. The cumulative F1 score over time for these configurations is given in the appendix, showing that although the impact of poor initialization is reduced over time, SECLEDS-rand does not completely recover from it. Thus, the distance-based non-uniform sampling strategy proves to be extremely helpful in initializing good clusters.

\paragraph{\textbf{Clustering with Concept drift. }}
We use Sine-curve-drifted with $k=3$, $p=5$, $\lambda=0.1$, $trials=10$ on stream sizes $n=(2000,\dots 30,000)$. 
SECLEDS outperforms \textit{all baselines} by 138.7\%, and outperforms the \textit{best-performing streaming baseline} by 143.7\%, on average. BanditPAM no longer serves as a benchmark because it only has a static view of the data, \ie it does not distinguish between class distributions at $T=t_x$ and $T=t_{x+y}$. 
Both SECLEDS and CluStream maintain their F1 scores with concept drift, but SECLEDS is 161.8\% better than CluStream.
StreamKM++ and Minibatch k-means observe a significant reduction in their performance. We hypothesize that it might be due to the lack of exponential decay in k-means, which limits the movement of the centroids towards newer data. This experiment provides strong evidence for SECLEDS' ability to handle concept drift with only \texttt{k}-fixed clusters. 

\paragraph{\textbf{Runtime analysis. }}
StreamKM++ and Minibatch k-means are among the fastest clustering algorithms on all datasets, which is expected since they are based on k-means. 
CluStream does not scale well for high-dimensional datasets, and is much slower than SECLEDS on sequence clustering. 
As the stream size $n$ grows, SECLEDS also becomes faster than the high-performance implementation of BanditPAM on both point and sequence clustering. 
Interestingly, the runtimes of BanditPAM, CluStream and StreamKM++ seem to be affected by concept drift: given the same dataset and constant parameters, their runtimes increase approximately twofold when there is drift in the data. We hypothesize that this is a side effect of the sampling strategy used to speed up these algorithms.

\begin{figure}[t]
    \centering
    \begin{subfigure}{\textwidth}
  \centering
\includegraphics[width=0.9\linewidth]{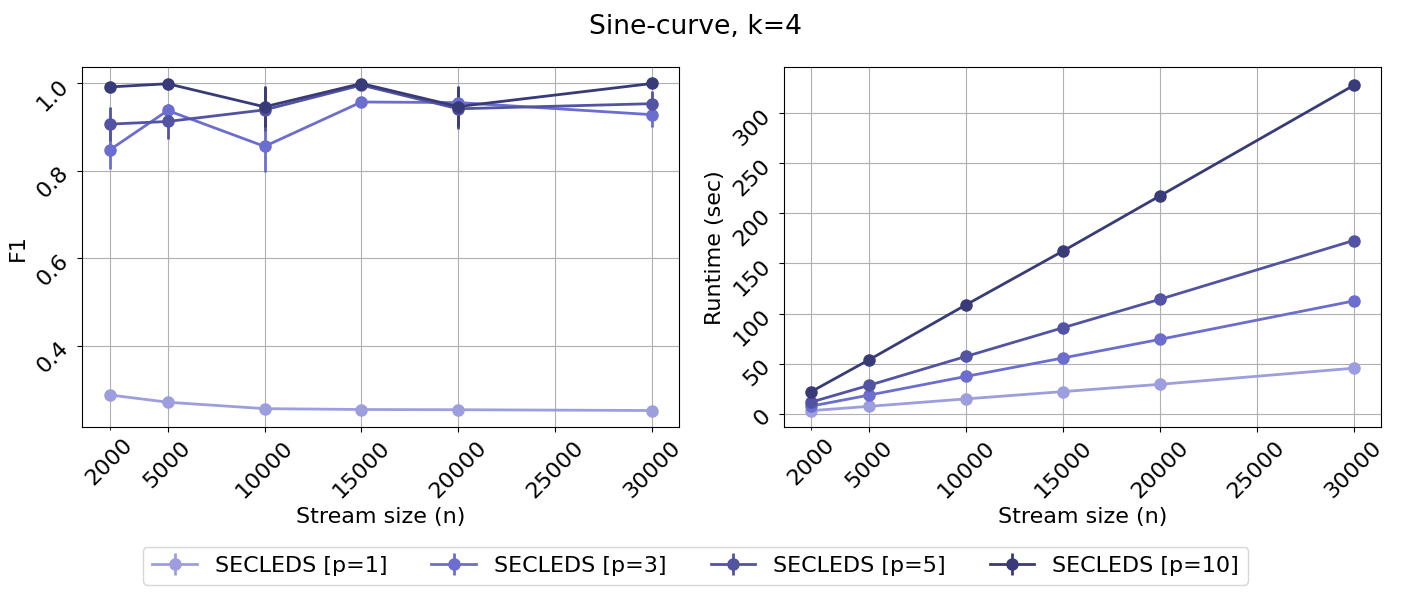}
    \caption{$p=\{3,5\}$ provide the best trade-off between runtime and F1 for Sine-curve.}
    \label{fig:experiments-scalingwithp}  %\caption{1a}
\end{subfigure}%

\begin{subfigure}{\textwidth}
  \centering
\includegraphics[width=0.9\linewidth]{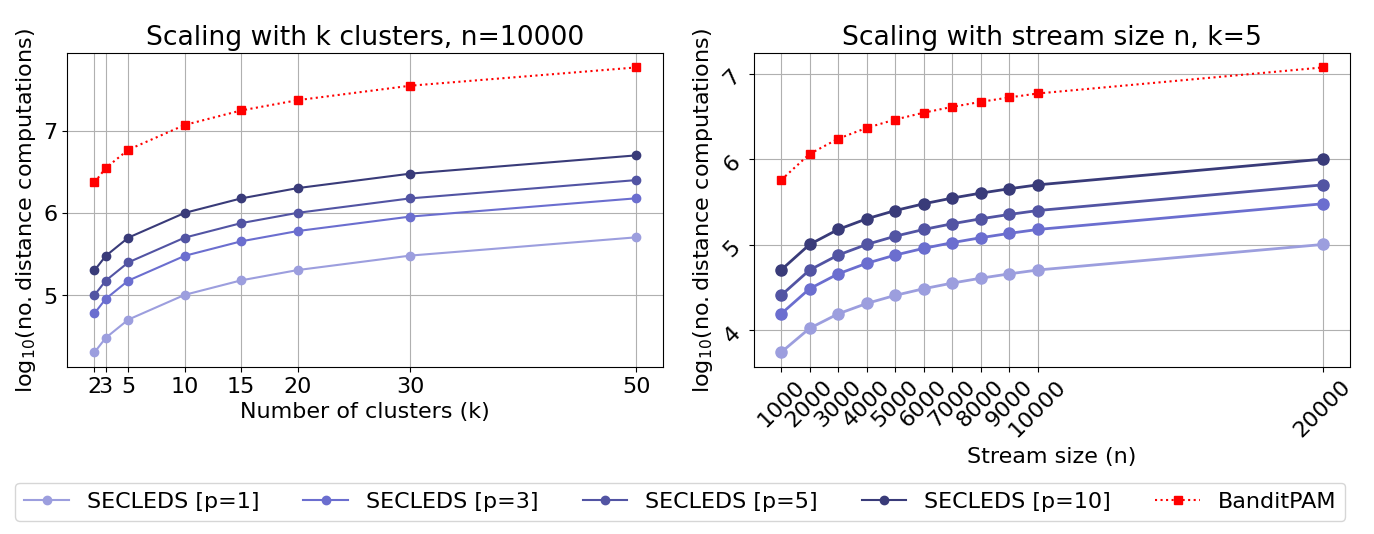}
    \caption{SECLEDS requires 83.7\% fewer distance computations compared to BanditPAM.}
\label{fig:dist-over-k}
\end{subfigure}%
\caption{Scaling with $n$, $k$, and $p$: (a) Empirical results; (b) Theoretical estimate.}
\label{fig:scalingwithp-overall}
    
\end{figure}

\paragraph{\textbf{Scaling with $n$, $k$, and $p$. }}
We use Sine-curve with $k=4$, $p=\{1,3,5,10\}$, $\lambda=0.1$, $trials=10$ on stream sizes $n=(2,000,\dots 30,000)$. The mean and standard deviation of the F1 and runtime of SECLEDS is reported in Figure \ref{fig:experiments-scalingwithp}. A single medoid per cluster, which is standard for PAM-based algorithms, does poorly in a streaming setting. Intuitively, more medoids help to improve the stability of the clusters, but the relationship is not linear. If \texttt{p} is set too low, the medoids keep jumping to various regions in the dataset, and if it is set too high, the medoids slow down the evolution of the clusters, having an equally detrimental effect on the performance. 
The optimal value of \texttt{p} with respect to performance and runtime is dataset-dependent. For Sine-curve, $p=\{3,5\}$ are good alternatives. 
Additionally, although SECLEDS has multiple medoids per cluster, it performs significantly fewer distance computations compared to the (almost linear) BanditPAM. Figure \ref{fig:dist-over-k} shows this for increasing stream size $n$, number of clusters $k$, and number of medoids $p$, with BanditPAM as reference.

\begin{table}[t]
\caption{Clustering real network traffic: Compared to BanditPAM, SECLEDS requires fewer distance computations, is faster, and has a better cluster quality. SECLEDS-dtw is slower but produces better clusters than SECLEDS. Overall, $\sim$5.5h of network traffic is clustered in under 27 minutes (Bold = best scores).}
\label{tab:netflow-results-k2dim100}
\centering
\resizebox{\columnwidth}{!}{%
\begin{tabular}{lccccc}
\hline
\textbf{} & \textbf{\begin{tabular}[c]{@{}c@{}}Stream \\ config.\end{tabular}} & \textbf{\begin{tabular}[c]{@{}c@{}}\# Distances \\ ($k=2$)\end{tabular}} & \textbf{\begin{tabular}[c]{@{}c@{}}Run time \\ ($k=2$)\end{tabular}} & \textbf{\begin{tabular}[c]{@{}c@{}}F1 \\ ($k=2$)\end{tabular}} & \textbf{\begin{tabular}[c]{@{}c@{}}F1 \\ ($k=5$)\end{tabular}} \\ \hline
\multirow{2}{*}{\textbf{BanditPAM}} & \begin{tabular}[c]{@{}c@{}}Time-ordered\end{tabular} & \multirow{2}{*}{$10.3\times 10^6$ } & 978.03s & 0.64 &  0.38 \\ \cline{2-2} \cline{4-6} 
 & \begin{tabular}[c]{@{}c@{}}Cross-validated\end{tabular} &  & 984.8s & 0.64 &  0.38 \\ \hline
\multirow{2}{*}{\textbf{SECLEDS}} & \begin{tabular}[c]{@{}c@{}}Time-ordered\end{tabular} & \multirow{2}{*}{$\mathbf{2.1\times 10^6}$} & \textbf{629.39s} & \textbf{0.85} & 0.82 \\ \cline{2-2} \cline{4-6} 
 & \begin{tabular}[c]{@{}c@{}}Cross-validated\end{tabular} &  & \textbf{631.84s} & 0.79 & 0.76\\ \hline
\multirow{2}{*}{\textbf{SECLEDS-dtw}} & \begin{tabular}[c]{@{}c@{}}Time-ordered\end{tabular} & \multirow{2}{*}{$\mathbf{2.1\times 10^6}$} & 1623.05s & \textbf{0.85} & \textbf{0.88} \\ \cline{2-2} \cline{4-6} 
 & \begin{tabular}[c]{@{}c@{}}Cross-validated\end{tabular} &  & 1626.89s & \textbf{0.81} & \textbf{0.80} \\ \hline
\end{tabular}}
\end{table}

\begin{figure}[t]
    \centering
    \begin{subfigure}{0.34\textwidth}
    \centering
    \includegraphics[width=\linewidth]{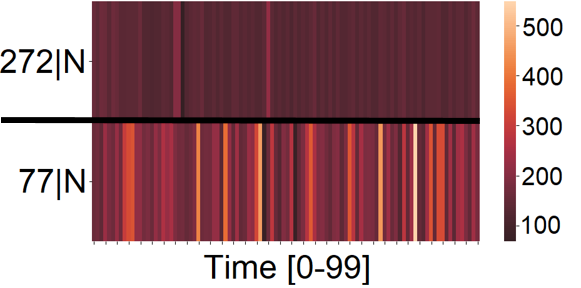}
    \caption{BanditPAM}
    \label{fig:bpamTH}
    \end{subfigure}%
    \begin{subfigure}{0.34\textwidth}
    \centering
    \includegraphics[width=\linewidth]{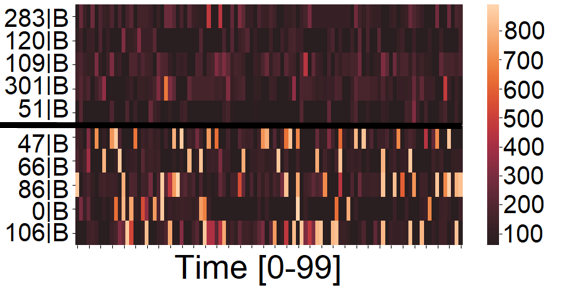}
    \caption{SECLEDS-dtw}
    \label{fig:secledsTH}
    \end{subfigure}%
    \begin{subfigure}{0.34\textwidth}
    \centering
    \includegraphics[width=\linewidth]{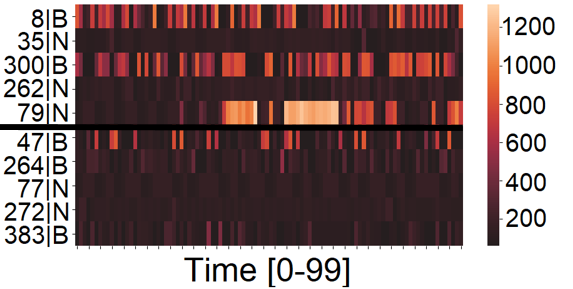}
    \caption{SECLEDS}
    \label{fig:secledsdtwTH}
    \end{subfigure}%
\caption{Visualizing the medoids of BanditPAM, SECLEDS \& SECLEDS-dtw on $k=2,\ p=5$. Each row is a medoid. The label denotes curve identifier and $y_i$.}
\label{fig:ctu-bpamsqclutemporalHM}
\end{figure}

\subsection{Use case: Intelligent network traffic sampling via SECLEDS}\label{sec:quant-study}

A typical enterprise network has a bandwidth of 25 Mbps\footnote{\url{https://mosaicnetworx.com/it-challenges/bits-bytes-understanding-enterprise-network-speeds/}}, which produces about 17,000 packets \textit{per second}, consuming 2 terabytes of storage space \textit{each day}! To conserve space, the packets are aggregated into network flows (netflows) at the router level, and only a fraction of them are stored for analysis \ie 1 in $N$ netflows are stored.  
Naturally, randomly sampled network traffic does not preserve the temporal patterns of the data, thus limiting the efficacy of traffic profiling and behavior analytics. 

We propose to cluster sequences of netflows using SECLEDS, and to periodically store a medoid snapshot of each cluster, since they are representative of the network traffic seen so far. 
This way, each snapshot stores an overview of temporally-linked netflows.
The number of clusters $k$ can be chosen depending on the required granularity of behaviors captured by the clusters. It can also be approximated from an initial batch using, \eg \cite{de2011extending}. The number of medoids $p$ can be configured according to available storage space, network bandwidth, and the intervals at which to store the medoids. 

We demonstrate this use case by generating a stream from the CTU13-9 netflows. The construction and feature engineering processes are given in the appendix. In short, the ground truth provides two classes, \ie $y_i \in \{botnet, normal\}$. Univariate sequences of average bytes per netflow are used to separate the two classes. We use two configurations for the stream: i) \textit{Time-ordered:} the sequences arrive in order of their timestamps; ii) \textit{Cross-validated:} we shuffle the stream. We run SECLEDS and SECLEDS-dtw with $k=2$, $p=5$, $trials=5$, and compare the performance against BanditPAM. The results are given in Table \ref{tab:netflow-results-k2dim100}.

SECLEDS is faster and produces better medoids compared to BanditPAM. Figure \ref{fig:ctu-bpamsqclutemporalHM} visualizes the final medoids produced by all three algorithms in the form of temporal heatmaps. Temporal heatmaps have previously been used to visualize temporal similarities in \cite{nadeem2021beyond}. Each row shows a sequence (medoid), and the colors indicate the magnitude of the curve at each time step. Both medoids of BanditPAM are from the normal class. Although the medoids of SECLEDS-dtw are all from the botnet class, it is evident that they capture distinct behaviors of the malicious hosts. SECLEDS finds medoids from both classes, but the clusters are impure, \ie more medoids on average are from different classes. As such, the cluster quality of SECLEDS-dtw is significantly better than that of SECLEDS.  

The results indicate that there are many smaller classes in the network stream, reflecting the various behaviors of benign and infected hosts. When $k$ is set to a larger number, the clustering algorithms find smaller, purer data regions, \eg for $k=5$, SECLEDS-dtw produces 4 pure clusters (2 normal and 2 botnet), while SECLEDS only produces 1 pure (normal) cluster, see appendix for their temporal heatmaps. Table \ref{tab:netflow-results-k2dim100} shows the F1 scores for $k=5$. Note that although the clustering results for $k=5$ are better than $k=2$, the former obtains a lower F1 score as a side-effect of the metric: it penalizes higher number of clusters when less class labels are available by lowering the recall. As such, we recommend to over-estimate $k$ in order to sample many regions from the network traffic. 

Finally, SECLEDS is faster than SECLEDS-dtw, as expected. SECLEDS clusters the entire stream in 3.1\% of the traffic collection time, and SECLEDS-dtw in 8\% of the collection time. This experiment provides evidence that SECLEDS can handle much larger network bandwidths.

\paragraph{\textbf{Network bandwidth support. }}
SECLEDS-dtw can handle networks with a bandwidth of up to 1.08 Gbps, which is more than sufficient for small to medium enterprises. The experiments in Table \ref{tab:netflow-results-k2dim100} show that SECLEDS spends $\frac{1626.89}{213386} = 0.0076$ seconds on average to cluster a single sequence of length $w=100$. Thus, SECLEDS can cluster $131.58$ sequences per second. Assuming that the sequence windows $w$ are non-overlapping over the traffic stream, SECLEDS can process $13,158$ individual netflows per second. The CTU13-9 dataset is composed of $115,415,321$ packets aggregated into $2,087,509$ netflows. Assuming uniform distribution, each netflow contains about $55.2$ packets. SECLEDS can, thus, process $726,315.79$ packets per second. Given that each network packet is about $1500$ bytes, this makes a total of $1.089$ Gigabytes per second. Similarly, SECLEDS can handle network bandwidths of up to 2.79 Gbps.

\section{Conclusions}
We propose SECLEDS, a streaming version of k-medoids with constant memory footprint.
SECLEDS uses a combination of multiple medoids per cluster and a medoid voting scheme to create \texttt{k}-clusters that evolve with evolving data streams.
Testing on several real and synthetic datasets and comparing against state-of-the-art baselines, we demonstrate that i) SECLEDS achieves competitive F1 score compared to the benchmark (BanditPAM) on streams without concept drift; ii) SECLEDS outperforms all baselines by 138.7\% on streams with concept drift; iii) SECLEDS reduces the number of required distance computations by 83.7\% compared to the benchmark, making it faster than BanditPAM and CluStream for several clustering tasks, iv) SECLEDS can support high-bandwidth network streams of up to 1.08 Gbps using the expensive dynamic time warping distance. 
These results reinforce the importance of designing lightweight medoid-based stream clustering algorithms.

\subsubsection{Acknowledgements} 
We thank Ruben te Wierik, Silviu Fucarev, and Rami Al-Obaidi for their contributions to the SECLEDS algorithm. 

\bibliographystyle{splncs04}

\appendix
\section{Dataset generation process}

\begin{figure}[b!]
\begin{subfigure}{.33\textwidth}
  \centering
  \includegraphics[width=\linewidth]{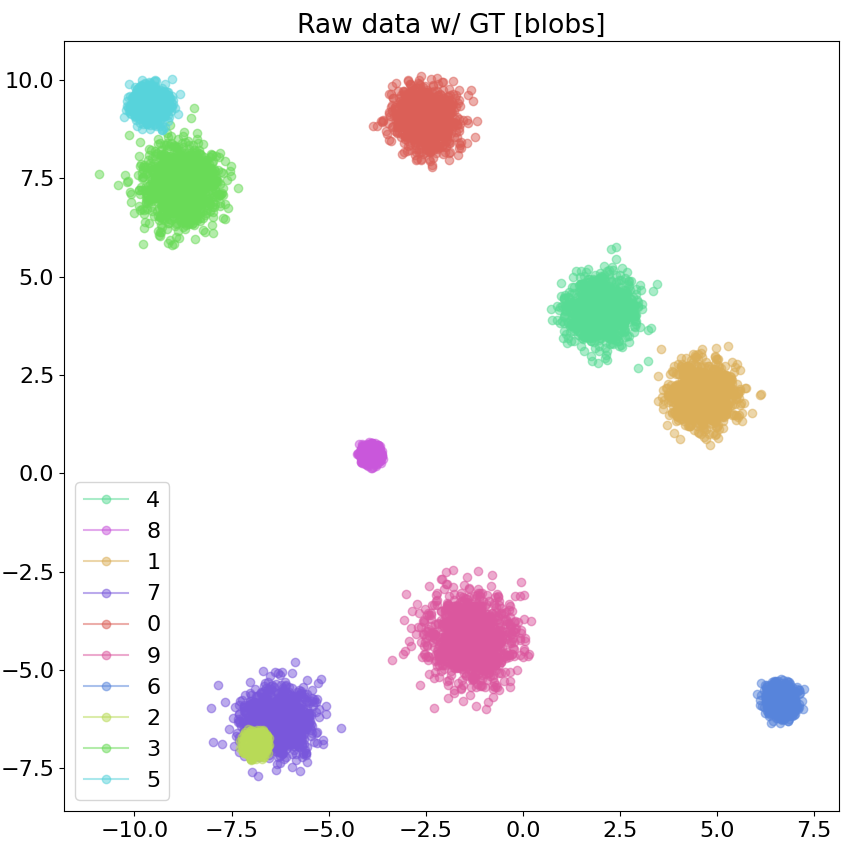}
\end{subfigure}%
\begin{subfigure}{.33\textwidth}
  \centering
  \includegraphics[width=\linewidth]{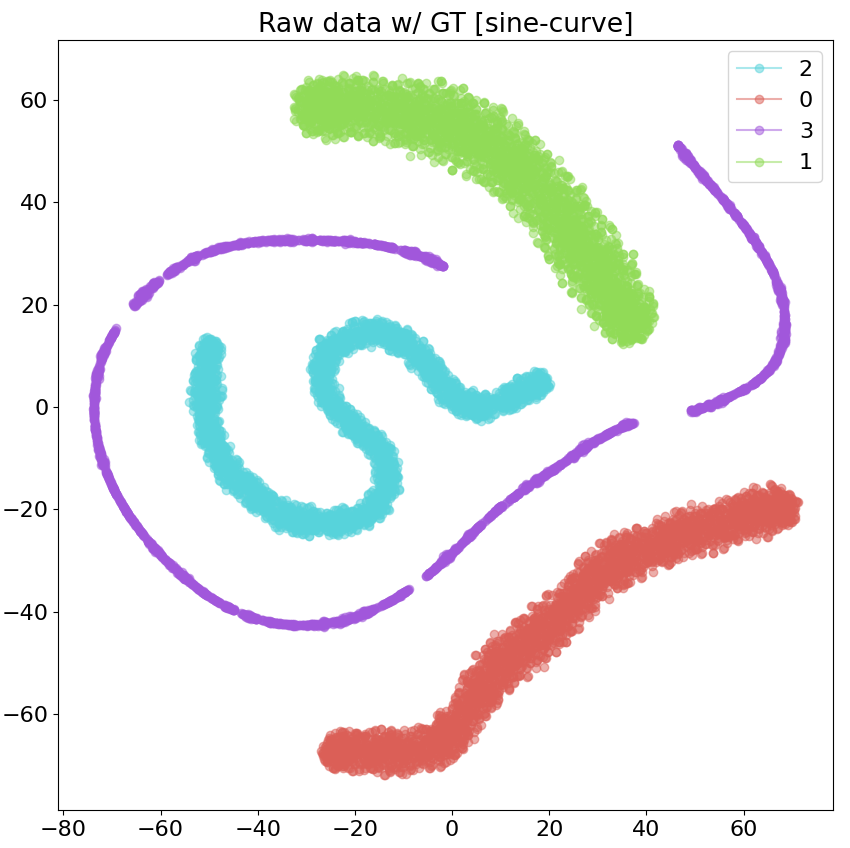}
\end{subfigure}
\begin{subfigure}{.33\textwidth}
  \centering
  \includegraphics[width=\linewidth]{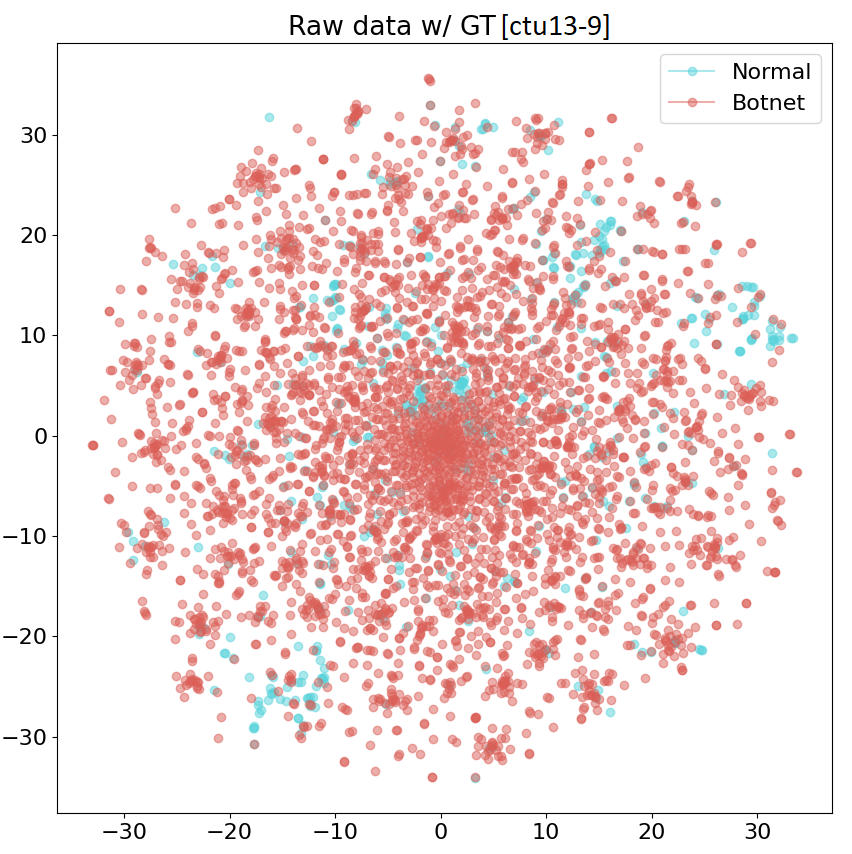}
\end{subfigure}

\begin{subfigure}{.33\textwidth}
  \centering
  \includegraphics[width=\linewidth]{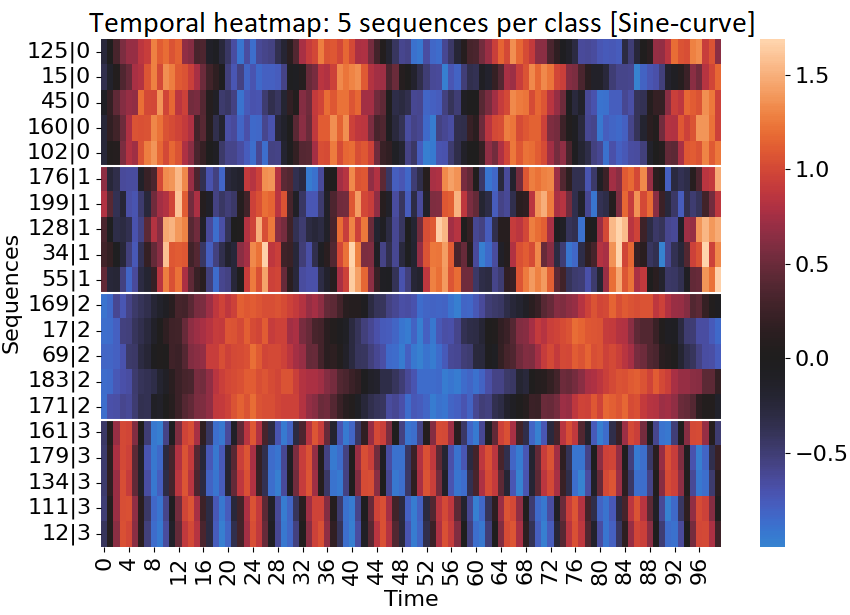}
\end{subfigure}
\begin{subfigure}{.33\textwidth}
  \centering
  \includegraphics[width=\linewidth]{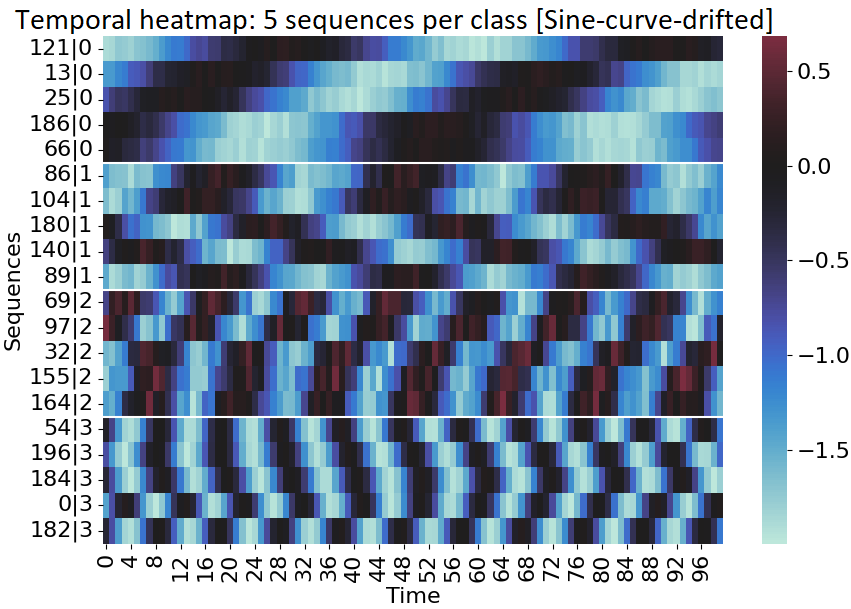}
\end{subfigure}
\begin{subfigure}{.32\textwidth}
  \centering
  \includegraphics[width=\linewidth]{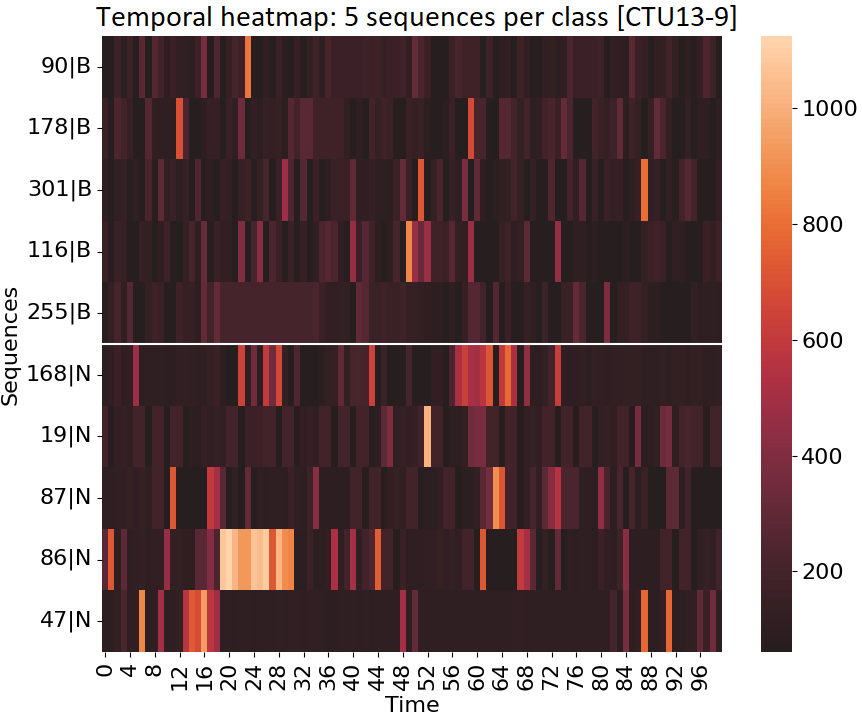}
\end{subfigure}
\caption{Experimental datasets: (Top left): Scatter plot showing the Blobs dataset ($k=10$). (Top middle): t-SNE plot showing a 2D representation of the Sine-curve dataset ($k=4$). (Top right): t-SNE plot showing a 2D representation of the CTU13-9 dataset ($k=2$). (Bottom left): Temporal heatmap showing 20 randomly sampled sequences from Sine-curve. (Bottom middle) Temporal heatmap showing 20 sequences from Sine-curve-drifted. (Bottom right): Temporal heatmap showing 10 sequences from CTU13-9.}
\label{fig:experimental-datasets-tsne}
\end{figure}

\textbf{Sine-curve. } The parameters used to create the 4 sine curves are: $frequency=\{(0.1, 0.12), (0.2, 0.22), (0.4, 0.42), (0.6, 0.62)\}$, $error=\{0.2, 0.4, 0.7, 0.1\}$ and $phase=\{5, 12, -10, -20\}$. For each curve $i \in \{1, \dots k\}$: (a) the phase value is $phase_i$, (b) a random frequency is selected from $frequency_i$ range, and (c) a random error of factor $error_i$ is added at every time step of the curve. Figure \ref{fig:experimental-datasets-tsne} shows the 2 dimensional representation of the dataset via a t-SNE plot and temporal heatmap. 5 randomly sampled sequences from the 4 classes are shown in the heatmap. The x-axis represents time, thus each row is a sequence. The colors show the magnitude of the curve at each time step. Temporal heatmaps have previously been used to visualize temporal similarities in (Nadeem,2021).

\textbf{CTU13-9. }  The network traffic is divided into several sub-streams based on the source IP address. There are 16 sub-streams, each collecting netflows for one of the 16 hosts in the dataset (10 infected with Neris botnet, 6 benign). We assume that the netflows arrive in sequences of length $w=100$, which are constructed by sliding a window $w$ over each of the sub-streams with $step size=1$. This results in $n=213,386$ netflow sequences. Using the two natural classes in the dataset, we follow (Garcia,2014) and label the sequences coming from infected sub-streams as \texttt{botnet}, while the others as \texttt{normal}. 

A preliminary feature engineering is conducted to find the netflow feature(s) that differentiates the two classes. There is extensive research studying the impact of various features in traffic classification. For simplicity, we consider a single feature at a time. Searching for better feature combinations is left as future work. We visualize the class separation obtained by i) netflow duration, ii) total bytes, iii) total packets, iv) inter-arrival time between netflows, v) average bytes per netflow, and vi) average packets per netflow. We select average bytes to separate the two classes. Figure \ref{fig:experimental-datasets-tsne} shows a t-SNE plot of the sequences. The unclear class separation emphasizes just how challenging the clustering task is.

\section{Cluster initialization quality}

\begin{figure}[t]
\centering
\begin{subfigure}{0.5\textwidth}
\centering
  \includegraphics[width=\linewidth]{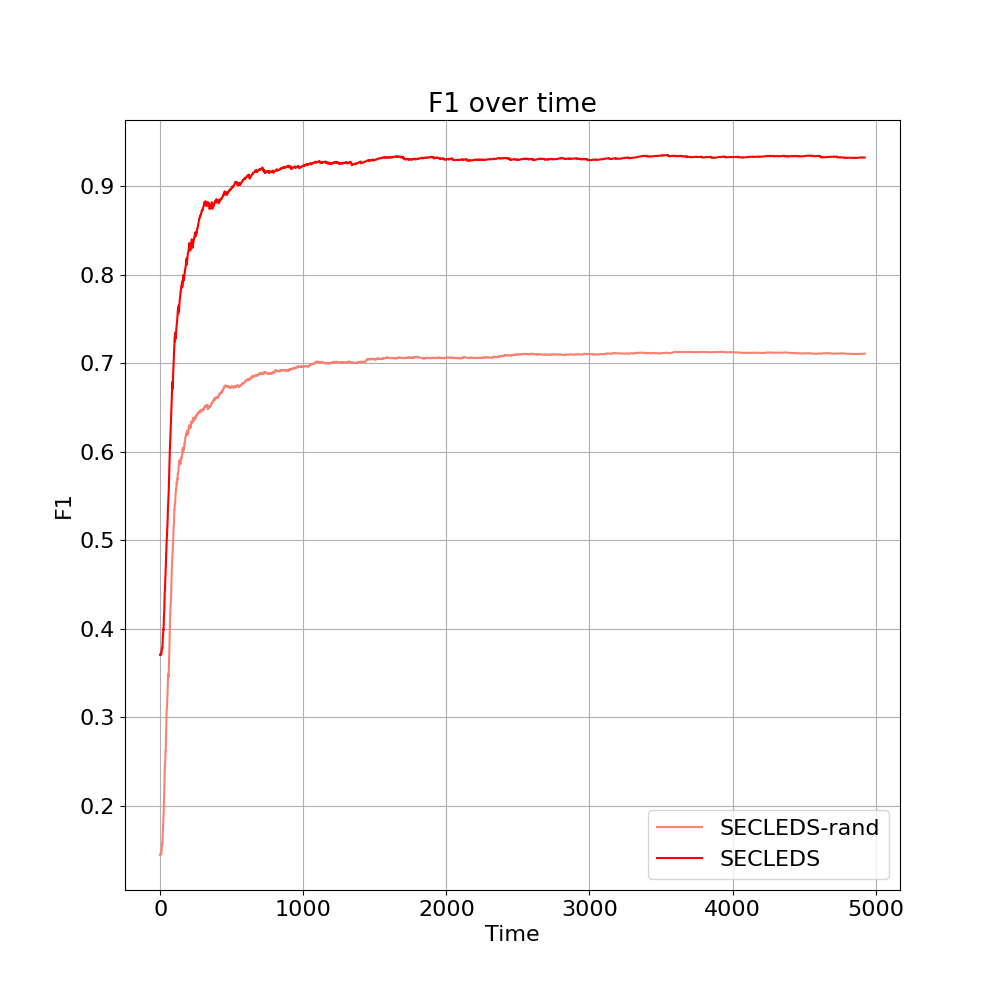}
  \caption{Blobs, $k=10;p=5;n=5,000$}
  %\label{fig:sfig1}
\end{subfigure}%
\begin{subfigure}{0.5\textwidth}
  \centering
  \includegraphics[width=\linewidth]{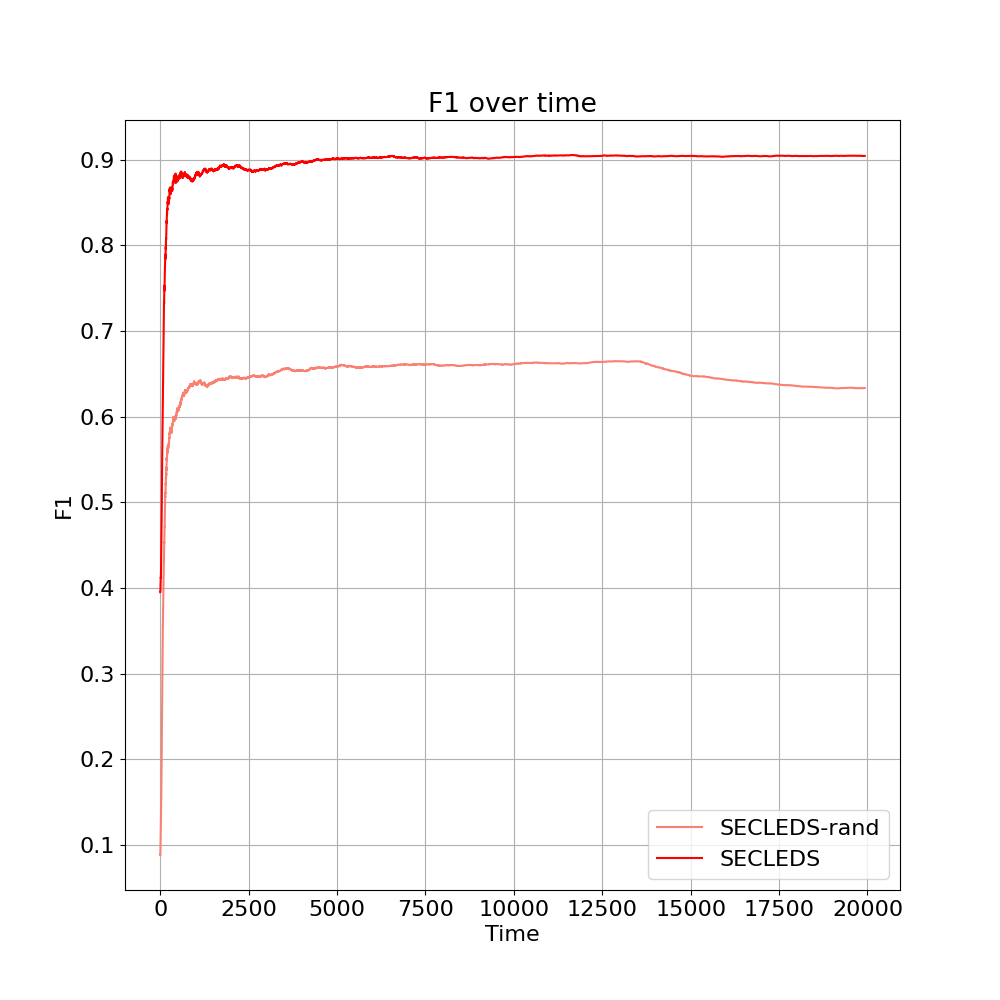}
  \caption{Blobs, $k=10;p=5;n=20,000$}
\end{subfigure}
\caption{F1 score over time for SECLEDS and SECLEDS-rand. Randomly initialized clusters perform sub-optimally regardless of stream size.}
\label{fig:effect-of-rand-init}
\end{figure}

Figure \ref{fig:effect-of-rand-init} shows the F1 scores over time for SECLEDS and SECLEDS-rand (randomly initialized medoids) for two stream sizes. It is evident that poorly initialized clusters are unable to perform optimally. Since the medoids in a single cluster are blind to each other, once poorly initialized, the clusters never converge. Further investigation is required to make the medoids aware of each other since the additional distance computations will increase the complexity of SECLEDS. 

\section{Network traffic sampling with SECLEDS}

\begin{figure}[t]
    \centering
    \begin{subfigure}{0.5\textwidth}
    \centering
    \includegraphics[width=\linewidth]{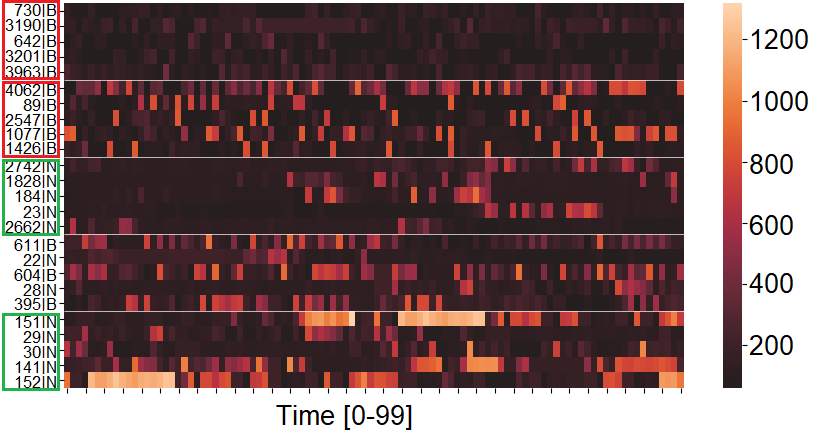}
    \caption{SECLEDS-dtw}
    \label{fig:secledsTH1}
    \end{subfigure}%
    \begin{subfigure}{0.5\textwidth}
    \centering
    \includegraphics[width=\linewidth]{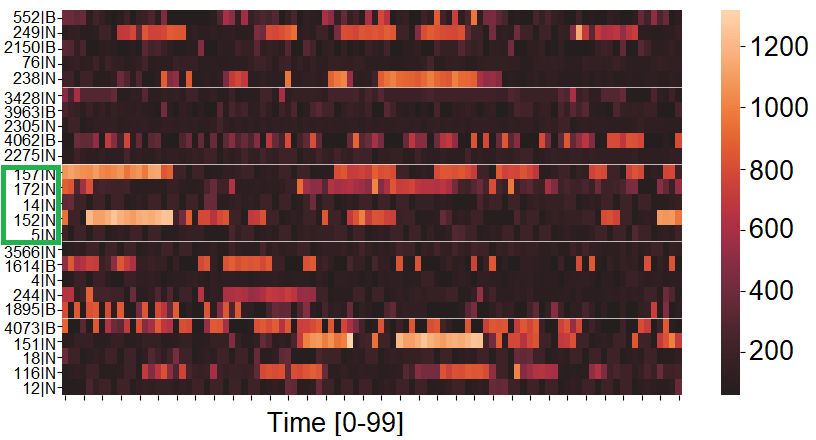}
    \caption{SECLEDS}
    \label{fig:secledsdtwTH1}
    \end{subfigure}%
\caption{Visualizing the medoids of SECLEDS-dtw \& SECLEDS on $k=5,\ p=5$. Each row is a medoid. The label denotes curve identifier and $y_i$. The pure clusters are highlighted by green for normal and red for botnet class.}
\label{fig:ctu-bpamsqclutemporalHMk5}
\end{figure}

The CTU13-9 dataset contains two big classes corresponding to the normal and botnet netflows. However, the experiments with $k=2$ indicate that there are several smaller regions in the data distribution, reflecting to the various behaviors of the normal and bot-infected hosts. We demonstrate this by setting $k=5$. Similar to the $k=2$ case, BanditPAM finds all clusters in the normal class. Figure \ref{fig:ctu-bpamsqclutemporalHMk5} shows the final medoids obtained by SECLEDS-dtw and SECLEDS. It shows that SECLEDS-dtw finds four fully pure clusters, \ie two in the botnet class and two in the normal class. SECLEDS (with Euclidean distance) struggles to find more than one pure cluster. This experiment provides evidence for alignment-based distance measures, such as dynamic time warping, to have superior performance than Euclidean distance for sequence clustering tasks. 

\section{Clustering Handwritten character strokes}
\begin{figure}[t]
    \centering
    \begin{subfigure}{0.5\textwidth}
    \centering
    \includegraphics[width=\linewidth]{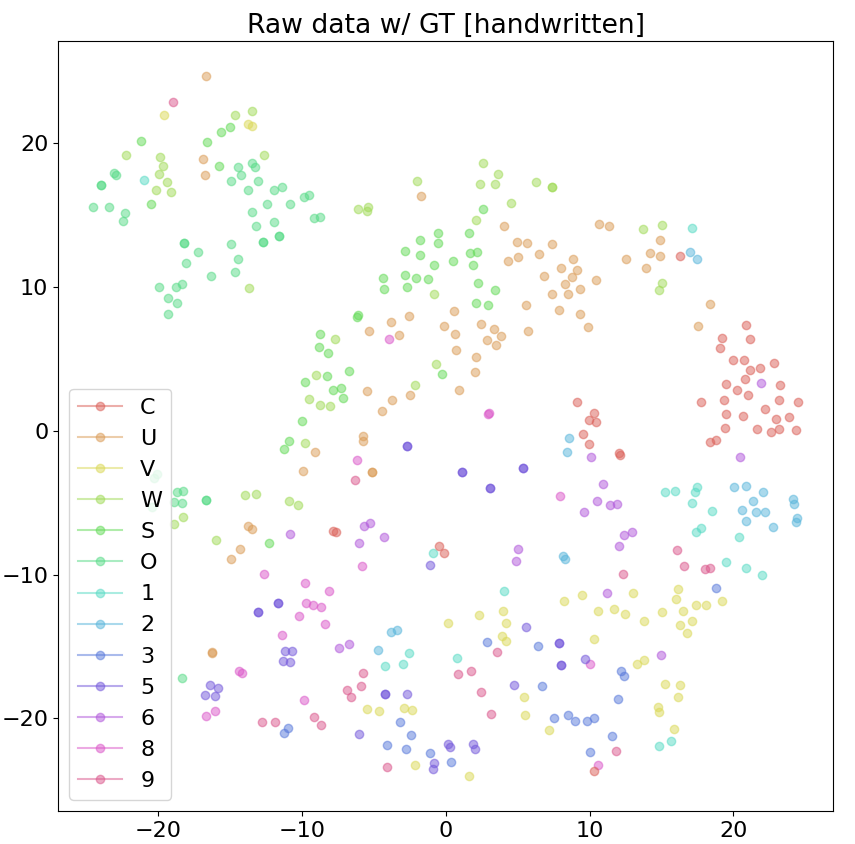}
    \end{subfigure}%
    \begin{subfigure}{0.5\textwidth}
    \centering
    \includegraphics[width=\linewidth]{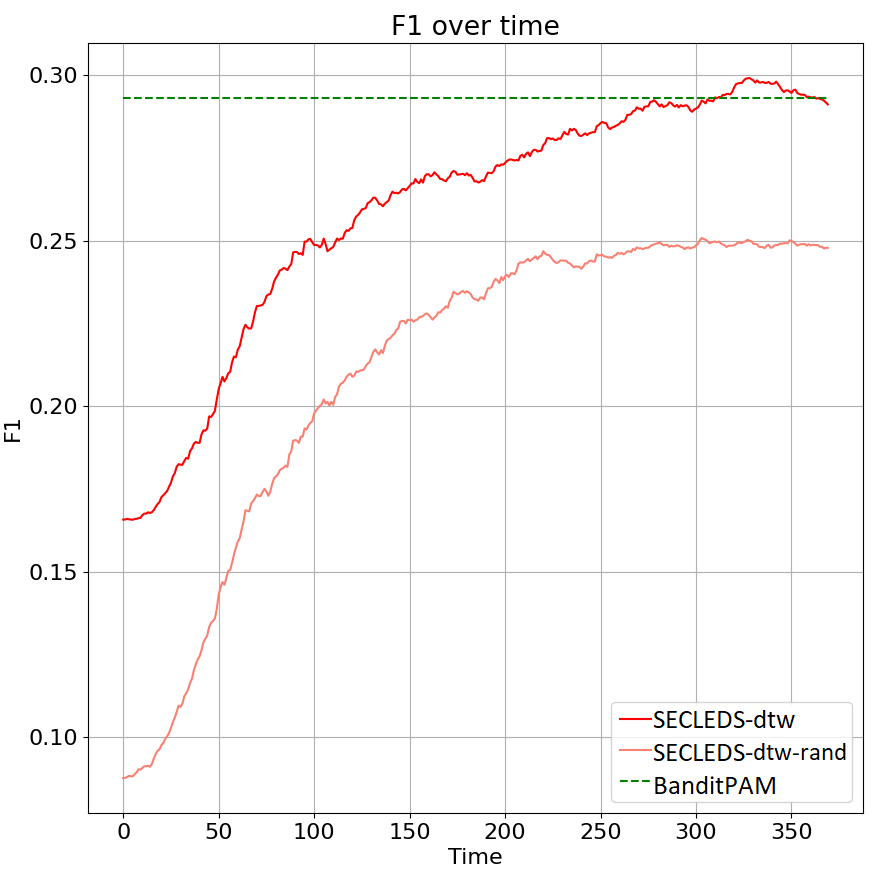}
    \end{subfigure}%
\caption{Clustering the \textit{UJI Pen Characters} dataset: (Left): t-SNE plot showing the 2D representation of the character strokes. (Right): F1 score over time for BanditPAM, SECLEDS-dtw and SECLEDS-dtw-rand. SECLEDS-dtw has comparable F1 score to BanditPAM, demonstrating that dynamic time warping is more suitable for sequence clustering in a streaming setting.}
\label{fig:handwritten-experiment}
\end{figure}

Despite the growing interest in sequential features over the years, one of the biggest bottlenecks in clustering sequences is the choice of an apt distance measure. Euclidean distance is by far the most popular metric used for clustering tasks. However, Euclidean distance does not perform accurately for out-of-sync sequential data, which is a common property of sequences. Although expensive, alignment-based distances, such as dynamic time warping, have instead been shown to be more accurate. Another benefit of dynamic time warping is that it can handle sequences of arbitrary lengths. We demonstrate this by clustering a dataset of handwritten characters from the UCI machine learning repository (Dua,2017). The \textit{UJI Pen Characters} dataset contains 467 bivariate sequences from 13 classes, \ie $\{C, O, S, U, V, W, 1, 2, 3, 5, 6, 8, 9\}$. The length of the shortest character stroke is 15, while that of the longest stroke is 121. Figure \ref{fig:handwritten-experiment} shows the t-SNE plot of the dataset.

SECLEDS-dtw is the only algorithm that can cluster this dataset straight out-of-the-box. The other baselines use Euclidean distance, which does not support arbitrary-length sequences. Figure \ref{fig:handwritten-experiment} also shows the 5-fold cross-validated F1 score over time for clustering the character strokes. BanditPAM is used as a reference benchmark, which requires the pairwise distance matrix of all sequences to cluster the dataset. Note that such a distance matrix is not available in a streaming setting, since it requires the entire dataset to be stored in memory. Additionally, we compare the performance with a randomly initialized SECLEDS, referred to as SECLEDS-dtw-rand. BanditPAM obtains the highest achievable F1 score of 0.294 since very few classes are properly separated. SECLEDS-dtw obtains a comparable F1 score of 0.291, while SECLEDS-dtw-rand obtains a score of 0.247. This experiment shows that dynamic time warping is indeed a more suitable distance measure for sequence clustering.

\end{document}